\pdfoutput=1

\documentclass[11pt]{article}

\usepackage[]{ACL2023}

\usepackage{times}
\usepackage{latexsym}

\usepackage[T1]{fontenc}

\usepackage[utf8]{inputenc}

\usepackage{microtype}

\usepackage{graphicx}
\usepackage{subfigure}
\usepackage{tabularx,booktabs}
\usepackage{enumitem}
\usepackage{amsmath}
\usepackage{inconsolata}
\usepackage{amssymb}
\usepackage{multirow}
\usepackage{algorithm,algorithmic}
\usepackage{pifont}
\usepackage{inconsolata}
\usepackage{colortbl}
\newcommand{\ouralg}{Recurrent-KIF} 

\title{Recurrent Knowledge Identification and Fusion for Language Model Continual Learning}

\author{Yujie Feng$^{1}$\thanks{~ Equal contribution.}\enspace, Xujia Wang$^{2}$\footnotemark[1]\enspace, Zexin Lu$^{1}$, Shenghong Fu$^{1}$, Guangyuan Shi$^{1}$ \\ \textbf{Yongxin Xu}$^{3}$\textbf{,} \textbf{Yasha Wang}$^{3}$\textbf{,} \textbf{Philip S. Yu}$^{4}$\textbf{,} \textbf{Xu Chu}$^{3}$\thanks{ ~ Corresponding author.}\enspace\textbf{,} \textbf{Xiao-Ming Wu}$^{1}$\footnotemark[2] \\
$^1$The Hong Kong Polytechnic University
$^2$Tsinghua University \\
$^3$Peking University 
$^4$University of Illinois at Chicago \\
 yujie.feng@connect.polyu.hk, xiao-ming.wu@polyu.edu.hk 
}

\begin{document}
\maketitle
\begin{abstract}
Continual learning (CL) is crucial for deploying large language models (LLMs) in dynamic real-world environments without costly retraining. 
While recent model ensemble and model merging methods guided by parameter importance have gained popularity, they often struggle to balance knowledge transfer and forgetting, mainly due to the reliance on static importance estimates during sequential training.
In this paper, we present {\ouralg}, a novel CL framework for Recurrent Knowledge Identification and Fusion, which enables dynamic estimation of parameter importance distributions to enhance knowledge transfer.
Inspired by human continual learning, {\ouralg} employs an inner loop that rapidly adapts to new tasks while identifying important parameters, coupled with an outer loop that globally manages the fusion of new and historical knowledge through redundant knowledge pruning and key knowledge merging.
These inner-outer loops iteratively perform multiple rounds of fusion, allowing {\ouralg} to leverage intermediate training information and adaptively adjust fusion strategies based on evolving importance distributions. 
Extensive experiments on two CL benchmarks with various model sizes (from 770M to 13B) demonstrate that {\ouralg} effectively mitigates catastrophic forgetting and enhances knowledge transfer.\footnote{\url{https://github.com/WoodScene/Recurrent_KIF}}

\end{abstract}

\section{Introduction}
Incorporating continual learning (CL) capability into large language models (LLMs) is essential for enabling them to acquire knowledge from diverse tasks sequentially, a critical requirement for adapting to ever-changing environments without extensive retraining \cite{wang2024comprehensive, jiang2024interpretable, yu2024recent, chang2024survey}.  
An effective CL system must address two key challenges: (1) Catastrophic Forgetting (CF)~\cite{mccloskey1989catastrophic}, where previously acquired knowledge is lost when learning new tasks, and (2) Knowledge Transfer (KT)~\cite{ke2021achieving}, which involves leveraging new, related tasks to improve performance on prior tasks, and vice versa.


\begin{figure}[t]
  \centering
  \includegraphics[width=1\linewidth]{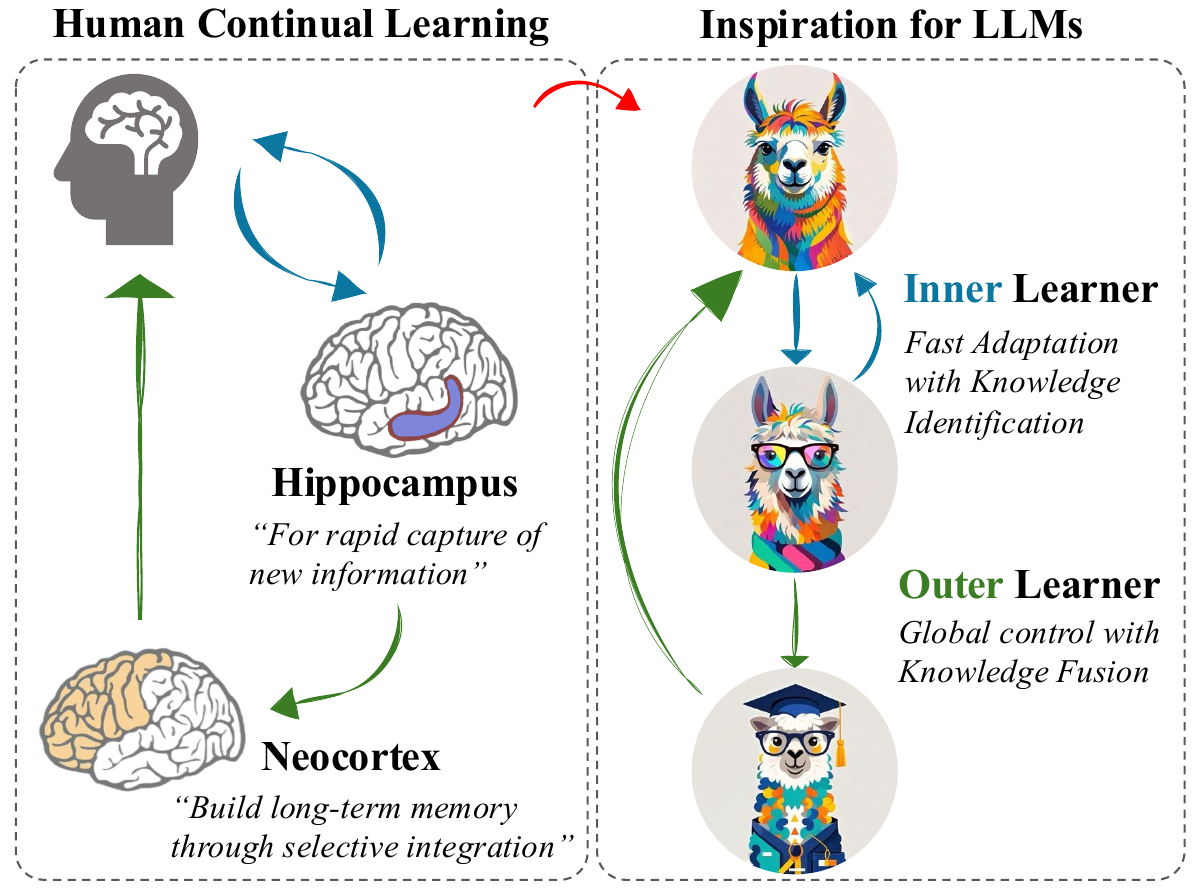}
  \caption{
  Conceptual illustration of {\ouralg}. Inspired by the CLS theory, {\ouralg} iteratively employs an inner learner to localize new knowledge and an outer learner to manage the global fusion of knowledge.
  }
  \label{fig:intro}
\end{figure}

Recently, model mixture-based methods have emerged as a mainstream approach for CL in LLMs \cite{chen2023lifelong, wu2024f, rypesc2024divide, chen2024entity}. 
By leveraging parameter-efficient fine-tuning (PEFT) techniques, which reduce the computational burden, these methods can be broadly classified into two categories: model ensemble and model merging.
Model ensemble methods assign a dedicated PEFT block to each task, capturing task-specific knowledge, which is then stored in a pool and dynamically selected during inference \cite{shengyuan2023differentiable, zhu2024llama, wang2024rehearsal}.
While effective, these methods require storing all task-specific models, leading to high memory consumption that grows with the number of tasks, which limits their scalability for long task sequences.

Another line of research focuses on model merging approaches \cite{dou-etal-2024-loramoe, wan2024knowledge, yadav2024survey}, which integrate new task knowledge after training into the historical model, maintaining a single unified model and reducing memory costs compared to model ensemble methods.
Consequently, our work primarily focuses on model merging approaches.
However, determining which parameters to merge and how to merge remains an open challenge \cite{qin2024large}.

Localizing important parameters in LLMs has recently gained significant interest, a topic widely explored in fields like model pruning and compression \cite{lu2021engage, panigrahi2023task, sun2023simple, yadav2024ties}. Building on this foundation, \citet{feng2024tasl} and \citet{du2024unlocking} have utilized gradient-based importance metrics, such as Hessian approximations, to identify critical parameters. By selectively or partially merging weights based on parameter importance, these methods have shown effectiveness in CL tasks.

However, the success of these approaches is contingent on the accurate estimation of parameter importance. A key limitation lies in their reliance on \textit{\textbf{static importance estimations}}, where the parameter importance scores for previous tasks remain unchanged and are not updated during subsequent training.
Over time, as model parameters gradually diverge from the state at which the Hessian was originally computed, these unadjusted importance estimates become increasingly inaccurate due to the growing truncation error in the Taylor expansion.
This issue is further detailed in the experiments section (Figure \ref{fig:visualization}).
The human brain demonstrates remarkable CL ability through two alternating systems: the hippocampus, which quickly acquires representations for specific experiences, and the neocortex, which selectively consolidates useful memories into long-term storage. 
This process is known as the Complementary Learning Systems (CLS) theory \cite{mcclelland1995there} in neuroscience.

Drawing inspiration from the CLS theory, we propose \textbf{Recurrent} \textbf{K}nowledge \textbf{I}dentification and \textbf{F}usion ({\ouralg}), a novel CL framework that dynamically estimates parameter importance and iteratively fuses knowledge.
{\ouralg} integrates an \textit{\textbf{inner learner}}, which rapidly adapts to new task-specific knowledge, and an \textit{\textbf{outer learner}}, which manages the global fusion of new and historical knowledge (see Figure \ref{fig:intro}).


In detail, the inner learner adapts to new knowledge while utilizing the proposed knowledge identification method to identify important parameters. The outer learner then retrieves historical task information from a memory buffer based on the latest model state, enabling dynamic updates of the importance distributions for previous tasks. Subsequently, a knowledge fusion mechanism is employed to integrate new and historical knowledge by pruning redundant information to mitigate CF and merging key knowledge to enhance KT.
Through iterative cycles of multiple rounds of fusion, {\ouralg} effectively captures valuable information throughout the model training process, distinguishing it from traditional post-training fusion methods. Each knowledge fusion step adaptively updates fusion weights according to the most recent importance distributions, resulting in smoother and more controlled optimization.

We conduct extensive experiments to assess the effectiveness of {\ouralg} on two CL benchmarks for LLMs. The results consistently highlight the superiority of {\ouralg} in mitigating CF while exhibiting exceptional KT capabilities, outperforming state-of-the-art methods.
Furthermore, {\ouralg} exhibits robust scalability across various model architectures and sizes (from 770M to 13B), underscoring its generalization ability.

Our main contributions are summarized as:
\begin{itemize}[leftmargin=*,itemsep=2pt,topsep=0pt,parsep=0pt]
\item 
We propose {\ouralg}, a novel CL \textbf{framework} for recurrent knowledge identification and fusion that dynamically estimates parameter importance and iteratively integrates knowledge.


\item 
We introduce a new \textbf{learning paradigm} for {\ouralg}, featuring an inner learner that rapidly captures and localizes new information, and an outer learner that globally controls the fusion of new and historical knowledge.

\item 


Extensive \textbf{evaluation} validates the effectiveness of {\ouralg} in addressing CL challenges.

\end{itemize}

\section{Related Work}
\subsection{Continual Learning for LLMs}
Continual learning (CL) \cite{zhou2024continual} focuses on developing algorithms that accumulate knowledge from non-stationary data. In the LLM era, model mixture-based methods using PEFT have become dominant \cite{wang2023rehearsal, huang2024mitigating, wang2024inscl}, typically divided into model ensemble and merging approaches.

Model ensemble methods isolate parameters by assigning independent PEFT blocks to each task \cite{feng2023towards, pham2023continual, ke2023sub, li2024revisiting, he2024seekr, wang2024self, zhang2025survey}. For example, O-LoRA \cite{wang2023orthogonal} enforces orthogonality among LoRA adapters, while SAPT \cite{zhao2024sapt} uses a selection module to combine blocks based on task correlations. 
While preserving task-specific knowledge, they hinder inter-task transfer and incur high memory overhead as the number of tasks increases, limiting their scalability.

In contrast, model merging methods combine multiple models into a single model \cite{cheng2024dam, alexandrov2024mitigating, ren2024analyzing, shengyuan2023differentiable, zhang2025survey}, alleviating memory constraints.
For example, global model merging approaches \cite{wortsman2022model, ilharco2023editing} perform a weighted fusion of models before and after training, typically assuming that all model weights contribute equally to each task.
However, determining which and how to merge parameters remains an open problem.
In this paper, we propose {\ouralg}, a novel framework that leverages the dynamic importance of parameters across different tasks by employing knowledge identification and fusion techniques to mitigate CF and promote KT.

\subsection{Parameter Importance Identification}
Identifying important parameters or knowledge regions within LLMs has gained significant attention in the NLP community \cite{zhao2023does, liu2023good, feng2024tasl2, xu2024parenting, shi2024understanding, zhang2025knowpo}. This research improves our understanding of LLMs and enhances their performance across a variety of tasks, including model editing \cite{wang2024editing, feng2025geoedit}, compression \cite{zhang2023adalora, jiang2023hykge}.

In the context of CL, \citet{du2024unlocking} use the gradient magnitudes to selectively update parameters. \citet{feng2024tasl} employ gradient-based metrics to compare the parameter importance distributions of current and historical tasks, merging task-shared regions to promote KT and retaining task-specific regions to prevent CF.
However, these approaches are limited by their reliance on static importance estimations for previous tasks, which become outdated as the model evolves. 

To address this limitation, \citet{wu2024meta} introduce VR-MCL, a replay-based method that dynamically updates importance information while reducing variance from random sampling. 
Although VR-MCL achieves dynamic importance estimation for historical tasks, it mainly focuses on preserving task-specific knowledge and does not update task-shared regions, thus limiting KT across tasks.
In contrast, inspired by the CLS theory, we propose a dynamic importance estimation method that iteratively updates parameter importance through inner and outer loops.
Our approach performs multi-round knowledge fusion, adaptively adjusting the integration of new and historical knowledge based on the latest model state. This method outperforms traditional post-training fusion by enhancing robustness and enabling smoother optimization.

\section{Proposed Method: {\ouralg}}

\begin{figure*}[t]
  \centering
  \includegraphics[width=1\linewidth]{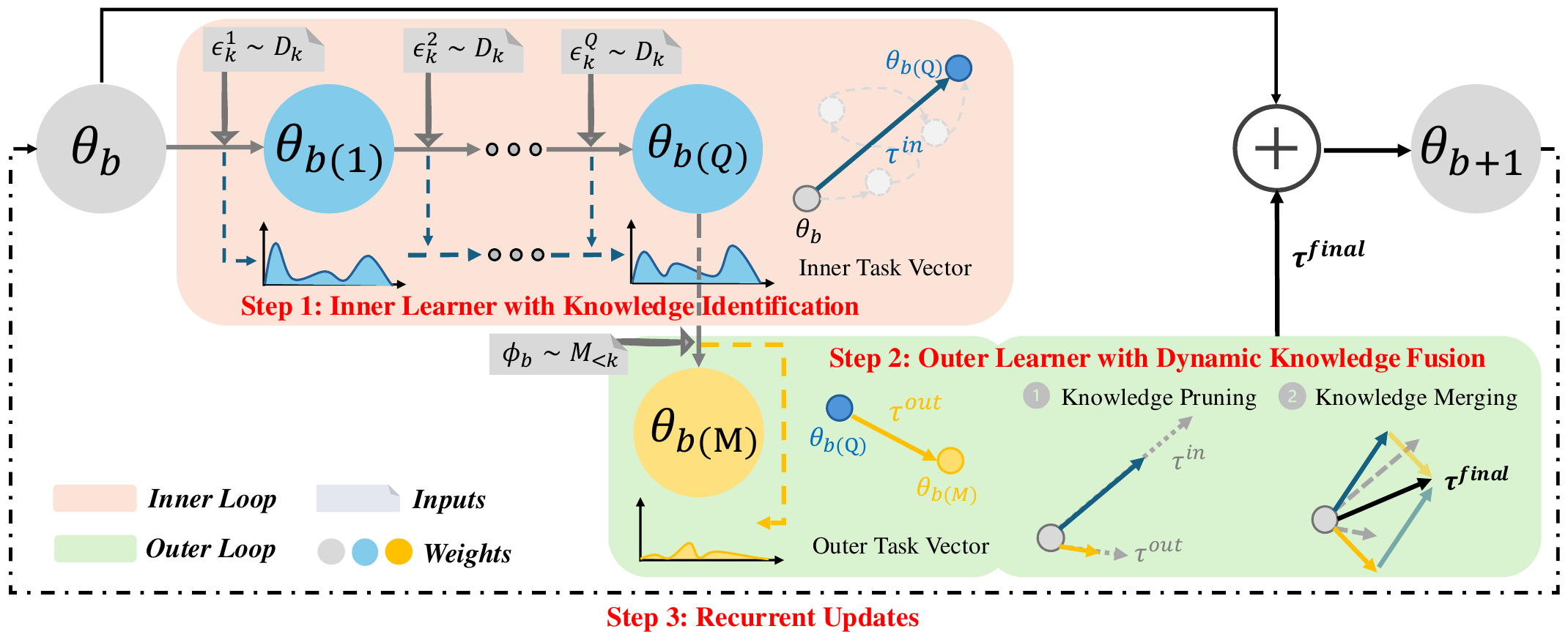}
  \caption{\textbf{Iterative update process of {\ouralg} for the $b$-th iteration.} 
  The notation $\epsilon_{k}^q$ represents training samples drawn from $\mathcal{D}_k$, while $\phi_{b}$ refers to samples drawn from $\mathcal{M}_{<k}$.
  \textbf{Inner Learner (Step 1):} Performs $Q$ iterations to rapidly adapt to the new task while identifying the parameter importance distribution.
    \textbf{Outer Learner (Step 2):} Retrieves historical task information using memory data and performs knowledge fusion, guided by the importance distributions of both current and historical tasks. 
    \textbf{Recurrent Updates (Step 3):} This inner-outer loop cycle is repeated, ensuring that each fusion knowledge step is based on up-to-date importance distributions.
  }
  \label{fig:method}
\end{figure*}

\paragraph{Problem Formulation}
Continual learning aims to progressively accumulate knowledge from a sequence of tasks $\{\mathcal{T}_1, \ldots, \mathcal{T}_K\}$. Each task $\mathcal{T}_k$ includes a distinct dataset $\mathcal{D}_k = \left\{ \left( x_i^k, y_i^k \right) \right\}_{i=1}^{N_k}$ of size $N_k$, where $x_i^k \in \mathcal{X}_k$ and $y_i^k \in \mathcal{Y}_k$.
The model, parameterized by $\Theta$, is trained sequentially on these tasks to minimize the following objective:
\begin{equation}
\mathcal{L} = \mathbb{E}_{(x, y) \sim \bigcup_{k=1}^K \mathcal{D}_k} \left[ -\log p_\Theta(y \mid x) \right]
\end{equation}

In this work, we consider a practical scenario where a small portion of data from previous tasks is stored in a memory buffer to facilitate the CL process. 
Specifically, we randomly store $\left| \mathcal{M} \right|$ samples from each task $\mathcal{T}_i$ in memory $\mathcal{M}_i$. During training, the model is jointly optimized on the new task data $\mathcal{D}_k$ and the memory buffer $\mathcal{M}_{<k}$.

\paragraph{Notation}
We consider a pre-trained model $\theta \in \mathbb{R}^n$ with $n$ parameters.
After training on task $\mathcal{T}_{k-1}$, the model are denoted as $\theta^{k-1}$.
Fine-tuning on a new task $\mathcal{T}_k$ produces updated parameters $\theta^k$.
The difference $\tau^k = \theta^k - \theta^{k-1}$, referred to as the \textit{task vector} or \textit{training residual} \cite{ilharco2023editing}, represents task-specific parameter updates.
In the {\ouralg} framework, we obtain transient training residuals through each iteration of the inner and outer loops. Specifically, two task vectors are employed to capture and quantify the new knowledge learned in the inner loop and the historical knowledge retrieved in the outer loop. 



\paragraph{Overview}
{\ouralg} restructures the training process into multiple iterative learning cycles, each comprising two key components as illustrated in Figure \ref{fig:method}:
(i) \textit{\textbf{Inner Learner with Knowledge Identification:}} rapidly acquires new task knowledge while estimating the corresponding parameter importance, and
(ii) \textit{\textbf{Outer Learner with Knowledge Fusion:}} utilizes a memory buffer to retrieve historical task information. 
By leveraging the importance distributions of both current and historical tasks, it provides global control for effective knowledge transfer through redundant knowledge pruning and key knowledge merging.


\subsection{Inner Learner with Knowledge Identification}  
Assume the current task is \(\mathcal{T}_k\), and the iterative update for the model parameters $\theta^{k-1}$ at the $b$-th iteration are denoted by $\theta_b^{k-1}$ \footnote{For simplicity, we omit the superscripts $k-1$ in subsequent descriptions.}.
In the inner loop, the model initializes with $\theta_{b(0)} = \theta_b$ and is rapidly updated over \(Q\) gradient steps using batch data $\epsilon_{k}^q$ sampled from \(\mathcal{D}_k\) at the $q$-th step. 
After obtaining $\theta_{b(Q)}$ the task-specific updates are encapsulated in the task vector $\tau_b^{in} \in \mathbb{R}^n$:
\begin{equation}
\tau_b^{\text{in}} = \theta_{b(Q)} - \theta_{b(0)}
\end{equation}

This task vector captures the knowledge acquired for the current task. However, $\tau^{in}$ often contains redundant information, and directly merging it into the model may compromise historical knowledge, leading to catastrophic forgetting.
To address this, we propose a knowledge identification technique to identify the key parameters which storing critical knowledge within the task vector.

We use a commonly adopted importance metric in model pruning \cite{konishi2023spg}, defined as the magnitude of the gradient-weight product:
\begin{equation}
\bar{I}\left(w_{i j}\right)=\left|w_{i j} \nabla_{w_{i j}} \mathcal{L}\right| \label{eq:1}
\end{equation}
where \(w_{ij}\) represents trainable parameters.  

Due to stochastic batch sampling and training dynamics, the metric in Eq. (\ref{eq:1}) may be unreliable, introducing variability \cite{zhang2022platon}. To mitigate this, we apply an exponential moving average \cite{zhang2023adalora} to smooth the trajectory gradients over $Q$ inner loop iterations:  
\begin{equation}
\begin{split}
I_{b(q)}   =\alpha_{1} I_{b(q-1)} + \left(1-\alpha_{1}\right) \bar{I}_{b(q)} \label{eq:I}
\end{split}
\end{equation}
where $\alpha_{1}$ is the smoothing factor, $q \in \left\{ 1, 2, ..., Q \right\}$ is the iteration number in the inner loop, and $I_{b(q)}$ represents smoothed importance.
The inner task vector \(\tau_b^{\text{in}}\) and its associated parameter importance \(I_b^{\text{in}}\) are then passed to the outer learner.



\subsection{Outer Learner with Knowledge Fusion}
The outer loop manages the global merging of knowledge, guided by parameter importance.
To access historical knowledge, after acquiring $\theta_{b(Q)}$, the outer loop samples data $\phi_{b}$ from the memory buffer $\mathcal{M}_{<k}$.
It then performs several training iterations, updating the parameters to $\theta_{b(M)}$. Then the outer task vector $\tau_b^{\text{out}} \in \mathbb{R}^n$, capturing historical task information, is defined as:
\begin{equation}
\tau_b^{\text{out}} = \theta_{b(M)} - \theta_{b(Q)}
\end{equation}

\paragraph{Dynamic Update of Historical Importance Distribution.}
While obtaining the outer task vector, we calculate the historical task importance distribution based on the latest model state $\theta_{b(Q)}$, using Eq. (\ref{eq:1}).
The update process is then expressed as:
\begin{equation}
\bar{I}_b^{\text{out}} = \mathbb{P}(\bar{I}_b^{\text{out}} \mid \theta_{b(Q)})
\end{equation} 

This update, based on conditional probability, enables the computation of the historical importance distribution $I_b^{\text{out}}$ using the current model state.
This distinguishes it from traditional static importance estimation methods and ensures more accurate knowledge identification. 
However, the limited sample size from the memory buffer can introduce significant variance in the importance estimates.
To address this, we also apply exponential smoothing to the previous outer loop distribution $I_{b-1}^{\text{out}}$:
\begin{equation}
I_b^{\text{out}} = \alpha_2 \bar{I}_b^{\text{out}} + (1 - \alpha_2) I_{b-1}^{\text{out}} \label{eq:out}
\end{equation} 
where $\alpha_2$ is the smoothing factor, enhancing stability and robustness in importance estimation.


\paragraph{Knowledge Fusion via Importance-based Binary Mask.}
Knowledge fusion is guided by the importance distributions $I_b^{\text{in}}$ and $I_b^{\text{out}}$.
To binarize the importance distributions, a quantile-based threshold $\delta$ is applied to select the top 20\% of parameters from both $I_b^{\text{in}}$ and $I_b^{\text{out}}$. This generates binary masks $m_b^{in} \in \mathbb{R}^n$ and $m_b^{out} \in \mathbb{R}^n$, defined as:
\begin{equation}
m_b^{\text{in}} = \mathbb{I}(I_b^{\text{in}} \geq \delta_b^{in}), m_b^{\text{out}} = \mathbb{I}(I_b^{\text{out}} \geq \delta_b^{out}) \label{eq:mask}
\end{equation} 
where $\mathbb{I}(\cdot)$ is the indicator function that outputs 1 if the condition is met and 0 otherwise.
Knowledge fusion is then performed as follows:
\begin{equation}
\theta_{b+1} = \theta_b + (m_b^{\text{in}} \odot \tau_b^{\text{in}} + m_b^{\text{out}} \odot \tau_b^{\text{out}}) \label{eq:fusion}
\end{equation} 
where $\odot$ denotes element-wise multiplication.

This knowledge fusion mechanism provides precise global control, effectively tackling key challenges in CL.
First, redundant information in the task vectors $\tau^{\text{in}}$ and $\tau^{\text{out}}$ is filtered out via the mask operation. Second, task-shared knowledge is effectively merged to facilitate knowledge transfer. Lastly, task-specific knowledge is preserved to prevent catastrophic forgetting.

The inner and outer loops operate iteratively, enabling multi-round fusion of knowledge. This iterative process facilitates the capture and absorption of useful information generated during training, providing smoother optimization compared to traditional post-training fusion methods.
Detailed implementation of {\ouralg} algorithm is provided in the Appendix (Algorithm~\ref{alg:my_algorithm}).


\section{Experiments and Analysis}\label{sec:exp}
\paragraph{Dataset} We adopt the experimental setup from \citet{du2024unlocking}, using two CL benchmark datasets:
(i) \textbf{Standard CL Benchmark}, which consists of five text classification tasks from \citet{zhang2015character}: AG News, Amazon Reviews, Yelp Reviews, DBpedia, and Yahoo Answers.
(ii) \textbf{Long Sequence Benchmark}, a more challenging evaluation scenario comprising 15 tasks \cite{razdaibiedina2023progressive}: five from the Standard CL Benchmark, four from the GLUE benchmark \cite{wang2018glue}, five from SuperGLUE \cite{wang2019superglue}, and the IMDB Movie Reviews dataset \cite{maas2011learning}.
Following \citet{wang2023orthogonal}, we sample 1000 instances for training on each task and reserve 500 per class for validation. Three task sequences are evaluated for each benchmark, with detailed descriptions and orderings provided in Appendix \ref{sec:dataset}.

\newcommand{\tabincell}[2]{\begin{tabular}{@{}#1@{}}#2\end{tabular}}
\begin{table*}[t]

\centering
\scalebox{0.85}{
\begin{tabular}{l|cccc}
\toprule
\multirow{2}*{\tabincell{c}{Method}}  & \multicolumn{2}{c}{Standard CL benchmarks} & \multicolumn{2}{c}{Long Sequence Benchmark} \\
  & OP$\uparrow$ &   BWT$\uparrow$ & OP$\uparrow$&   BWT$\uparrow$ \\
\midrule
\rule{0pt}{4pt}SeqLoRA  & 43.7 & -50.4 & 11.6 & -73.4 \\
\rule{0pt}{8pt}IncLoRA  & 66.4 & -20.0 & 61.2 & -26.7 \\
\rule{0pt}{8pt}LoRAReplay  & 68.8 & -11.7 & 70.9 & -15.4 \\
\rule{0pt}{8pt}EWC$^*$~\cite{kirkpatrick2017overcoming} & 50.3 & - & 45.1 & - \\
\rule{0pt}{8pt}L2P$^*$~\cite{wang2022learning}  & 60.7 & - & 56.1 &  -16.3\\
\rule{0pt}{8pt}LFPT5$^*$~\cite{qin2021lfpt5}  & 72.7 & - & 69.2 & -12.8 \\
\rule{0pt}{8pt}MoELoRA$^*$~\cite{luo2024moelora} & 54.1 & - & 27.6 & - \\
\rule{0pt}{8pt}O-LoRA$^*$~\cite{wang2023orthogonal} & 75.8 & -3.8  & 69.6 & -4.1 \\
\rule{0pt}{8pt}TaSL~\cite{feng2024tasl} & 76.3 & -4.0 & 74.4 & -5.3  \\
\rule{0pt}{8pt}VR-MCL~\cite{wu2024meta}  & 76.0 &  -3.7 & 74.8 & -4.9\\
\rule{0pt}{8pt}MIGU$^*$~\cite{du2024unlocking} & 76.6 & - & 76.5 & -   \\

\rowcolor[gray]{0.9}
\rule{0pt}{8pt}\textbf{{\ouralg} (ours)} & \textbf{78.4} & \textbf{-2.8} & \textbf{77.8} & \textbf{-3.6} \\

\midrule

\rule{0pt}{8pt} MTL &  80.3 & - & 81.8 & -  \\
\rule{0pt}{8pt} SAPT-LoRA~\cite{zhao2024sapt} & - & - & 82.0 & -1.3  \\

\bottomrule
\end{tabular}}
\caption{Overall results on two CL benchmarks using the T5-large model. We report Overall Performance (OP) and Backward Transfer (BWT) after training on the final task. All results are averaged over three different task orders. Methods marked with $*$ are copied from previous papers. The last two rows represent upper bound performance.
}
\label{tbl:result}
\end{table*}

\paragraph{Metrics}
Let $a_{i,j}$ denote the testing performance on task $\mathcal{T}_i$ after training on task $\mathcal{T}_j$.
We evaluate the overall performance (OP) \cite{chaudhry2018riemannian} and backward transfer (BWT) \cite{lopez2017gradient} after training on the final task:
\begin{equation}
    \mathbf{OP} =\frac{1}{K} \sum_{i=1}^{K} a_{i, K}
\end{equation}
\begin{equation}
    \mathbf{BWT} = \frac{1}{K-1} \sum\limits_{i=1}^{K-1} (a_{i, K}-a_{i, i})
\end{equation}

\paragraph{Baselines}
We compare {\ouralg} against a range of baseline methods, including traditional CL approaches, recent PEFT-based model ensemble and merging methods. 
(1) \textit{\textbf{SeqLoRA}:} LoRA parameters are trained on a task sequence without regularization or sample replay.
(2) \textit{\textbf{IncLoRA}:} incremental learning of LoRA parameters without regularization or sample replay.
(3) \textit{\textbf{LoRAReplay}:} LoRA fine-tuning with a memory buffer.
(4) \textit{\textbf{EWC \cite{kirkpatrick2017overcoming}}:} finetune LoRA with a regularization loss to prevent interference with previous tasks.
(5) \textit{\textbf{L2P \cite{wang2022learning}}:} dynamically selects and updates prompts from a pool on an instance-by-instance basis.
(6) \textit{\textbf{LFPT5 \cite{qin2021lfpt5}}:} learns a soft prompt that solves tasks and generates training samples for replay.
(7) \textit{\textbf{O-LoRA \cite{wang2023orthogonal}}:} extends IncLoRA to
learn different LoRAs in orthogonal subspaces.
(8) \textit{\textbf{MoELoRA \cite{luo2024moelora}}:} a vanilla MoE with LoRA number equals to the task number.
(9) \textit{\textbf{SAPT \cite{zhao2024sapt}}:} uses pseudo samples and a shared attention framework to align PEFT block learning and selection
(10) \textit{\textbf{TaSL \cite{feng2024tasl}}:} selectively updates or retains skill regions based on parameter importance.
(11) \textit{\textbf{MIGU \cite{du2024unlocking}}:} updates important parameters based on gradient magnitude.
(12) \textit{\textbf{VR-MCL \cite{wu2024meta}}:} dynamically updates historical task parameter importance distributions using memory replay.
Additionally, multi-task learning with LoRA, referred to as \textit{\textbf{MTL}}, serves as the upper bound.

\paragraph{Training Details}
We evaluate {\ouralg} using two distinct language model architectures: the encoder-decoder T5 model \cite{raffel2020exploring} (T5-large and T5-xl), and the decoder-only LLaMA model \cite{touvron2023llama} (LLaMA2-7B and LLaMA2-13B)\footnote{Due to being limited by an academic computing budget, we employed 8-bit quantization for LLaMA models.}.
Hyperparameters $\alpha_1$ and $\alpha_2$ in Eq. (\ref{eq:I}) and Eq. (\ref{eq:out}) are set to 0.55, with the number of inner loop iterations $Q$ set to 8, and the number of outer loop iterations set to 4.
Following \citet{zhao2024sapt}, 2\% of the original training set is used for replay samples.
All experiments are averaged over 3 runs. 
More details are in Appendix \ref{sec:details}.



\subsection{Main Results}
The overall CL results using the same T5-large backbone are summarized in Table \ref{tbl:result}.

\paragraph{Our {\ouralg} effectively addresses the challenges of CF and KT simultaneously.}
Compared to both traditional CL methods (LoRAReplay, L2P) and model ensemble-based methods (MoELoRA, O-LoRA), {\ouralg} outperforms them in both CF (increasing average OP from 72.7\% to 78.1\% compared to O-LoRA) and KT (increasing average BWT from -13.6\% to -3.2\% compared to LoRAReplay).
SAPT achieves the highest performance by leveraging generative replay-based data augmentation, surpassing MTL result. However, it relies heavily on external data synthesis, which can be costly in LLM settings.

Moreover, when compared to parameter importance-based methods like TaSL and VR-MCL, {\ouralg} consistently delivers the best OP and BWT scores. 
Notably, {\ouralg} outperforms the state-of-the-art CL method, MIGU, increasing OP from 76.6\% to 78.1\%.
These results underscore the effectiveness of our recurrent knowledge identification and fusion framework, validating the advantages of dynamic parameter importance estimation in mitigating forgetting and promoting knowledge transfer.

\begin{figure}[t]
  \centering
  \includegraphics[width=1\linewidth]{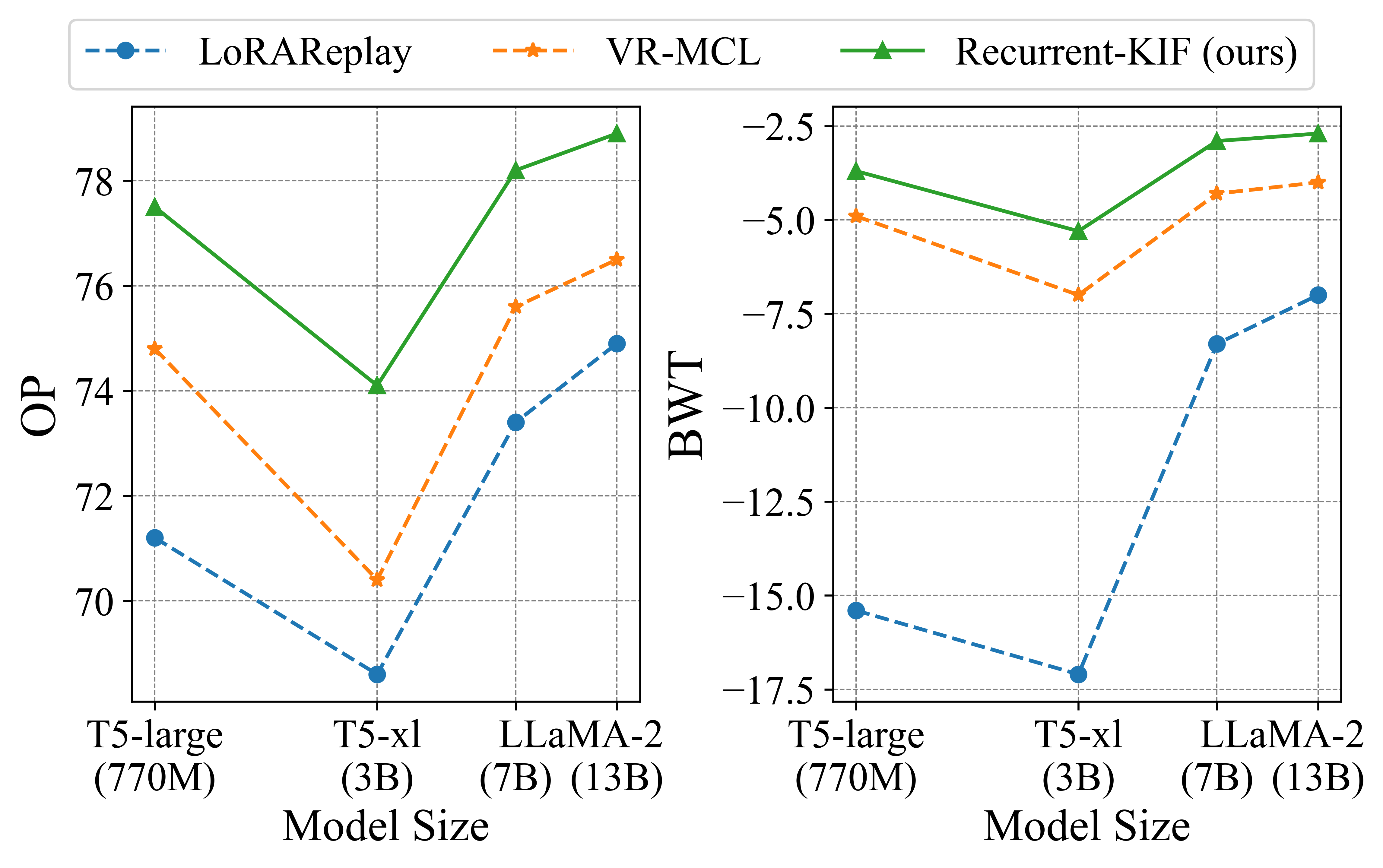}
  \caption{Performance of {\ouralg} with different backbones on the Long Sequence Benchmark.}
  \label{fig:different_size}
\end{figure}

\paragraph{{\ouralg} demonstrates consistent superiority across various backbones.}
To further validate the robustness of {\ouralg}, we conduct experiments across different backbones (Figure \ref{fig:different_size}). Across all backbone sizes, from 770M to 13B, {\ouralg} consistently outperforms all baseline models.
For instance, using the LLaMA2-7B backbone, {\ouralg} boosts the OP metric from 75.6\% to 78.2\% compared to VR-MCL. 
These results emphasize the critical role of accurate parameter importance estimation and demonstrate the robust generalization capability of {\ouralg} across different model scales.

\begin{figure}[t]
  \centering
  \includegraphics[width=1\linewidth]{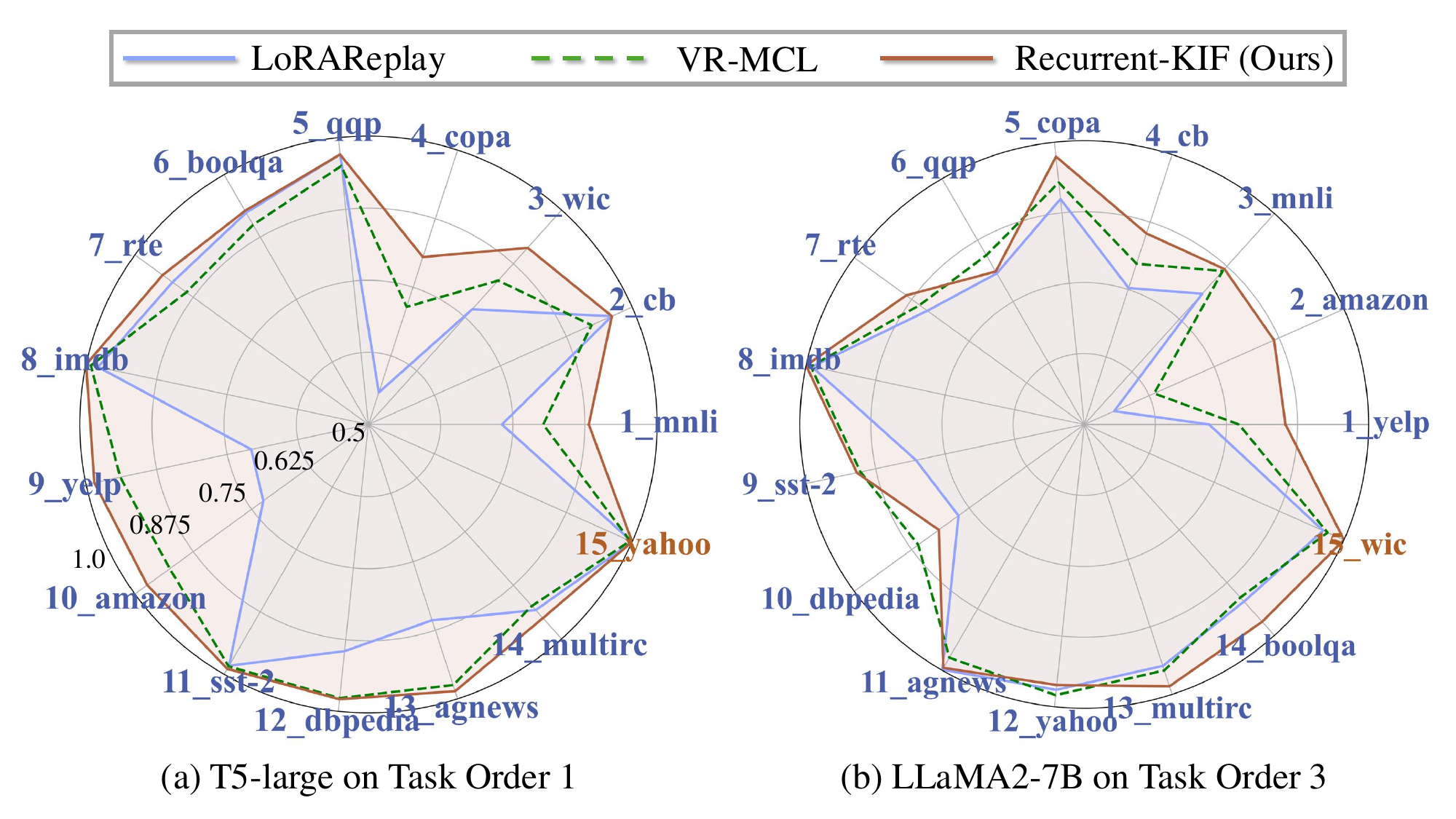}
  \caption{\textbf{Impact of Catastrophic Forgetting in Continual Learning.}
  After fine-tuning on the final task (in orange), {\ouralg} demonstrates superior resistance to performance decline on previously learned tasks (in blue), outperforming baseline methods.
  }
  \label{fig:radial_plot}
\end{figure}

\paragraph{Dynamic importance estimation enables effective knowledge retention and transfer.}
Figure \ref{fig:radial_plot} illustrates the performance of various methods across all historical tasks after completing the final task.
It highlights that {\ouralg} optimally restores the model’s performance on previous tasks, with significant improvements on Amazon and Copa.
Notably, on IMDB and AG News, {\ouralg} performs comparably to multi-task learning results. These demonstrate that {\ouralg} strikes an effective balance between preserving prior knowledge and excelling in new tasks.

\begin{table}[t]
\centering
\scalebox{0.9}{
\begin{tabular}{lcc}
\toprule
Method & OP &  BWT\\
\midrule
\rowcolor[gray]{1}
\rule{0pt}{6pt} {\ouralg}  & \textbf{77.9} & \textbf{-3.4} \\
\midrule
\rule{0pt}{8pt}  - DIE  & 74.8 & -4.8\\
\rule{0pt}{8pt}  - KI & 52.3 & -21.5  \\
\rule{0pt}{8pt} + GM & 72.1 & -11.2 \\
\rule{0pt}{8pt} + Adaptive & 76.1 & -4.1\\
\rule{0pt}{8pt} - Share & 75.8 &  -4.3\\

\bottomrule
\end{tabular}}
\caption{Ablation study. ``- DIE'', ``- KI'', ``- Share'' refer to the removal of dynamic importance estimation, knowledge identification, and task-shared region updates, respectively. ``+ GM'', ``+ Adaptive'' represent replacing the knowledge fusion mechanism with global merging and adaptive merging strategy, respectively.
}
\label{tbl:ablation}
\end{table}

\begin{figure*}[t]
  \centering
  \subfigure[Normalized inner task vector and corresponding parameter importance distribution for the current task.]{\includegraphics[width=0.45\linewidth]{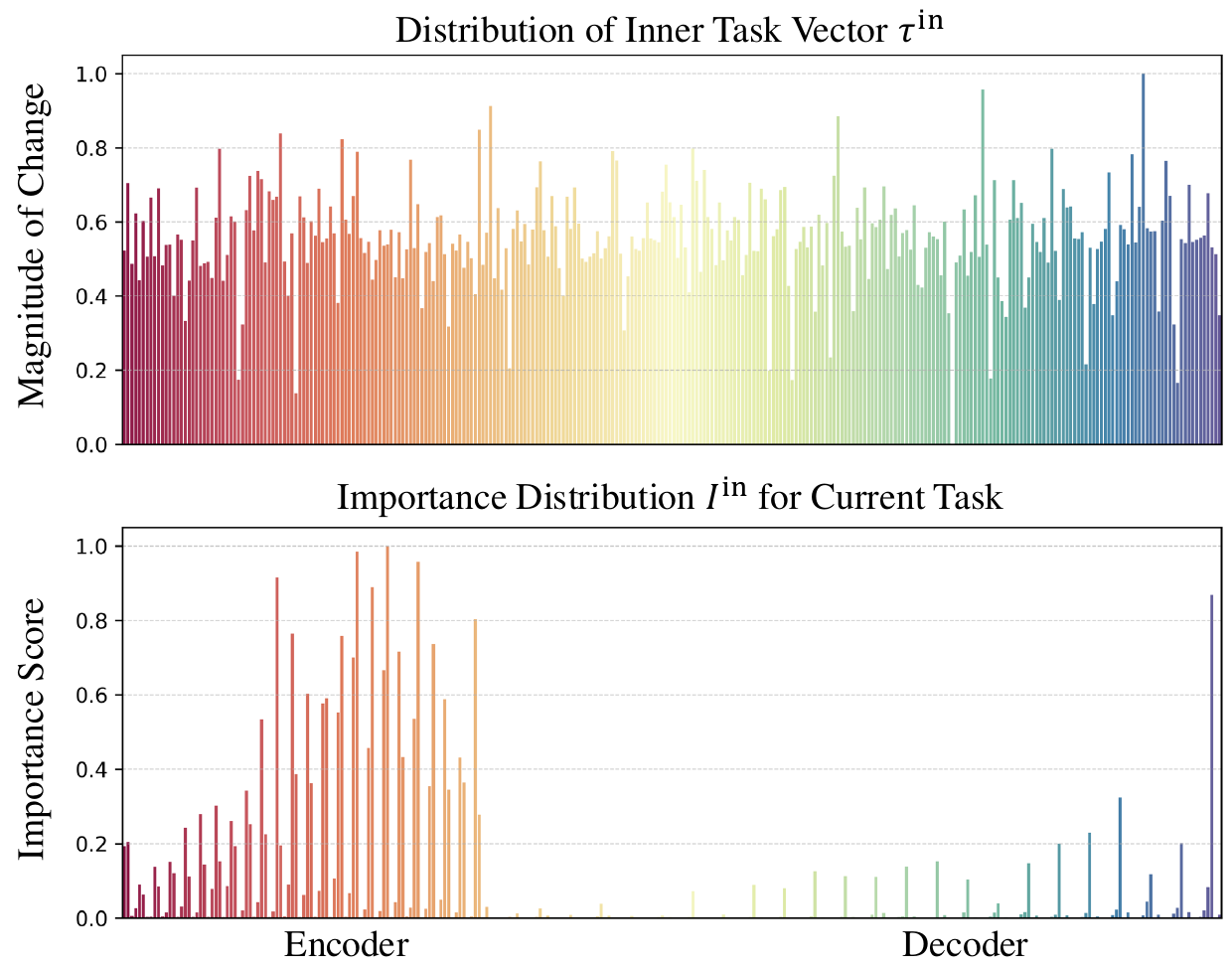}}
  \hspace{0.03\linewidth}
  \subfigure[Normalized parameter importance scores for a historical task across model states. 
  ]
  {\includegraphics[width=0.45\linewidth]{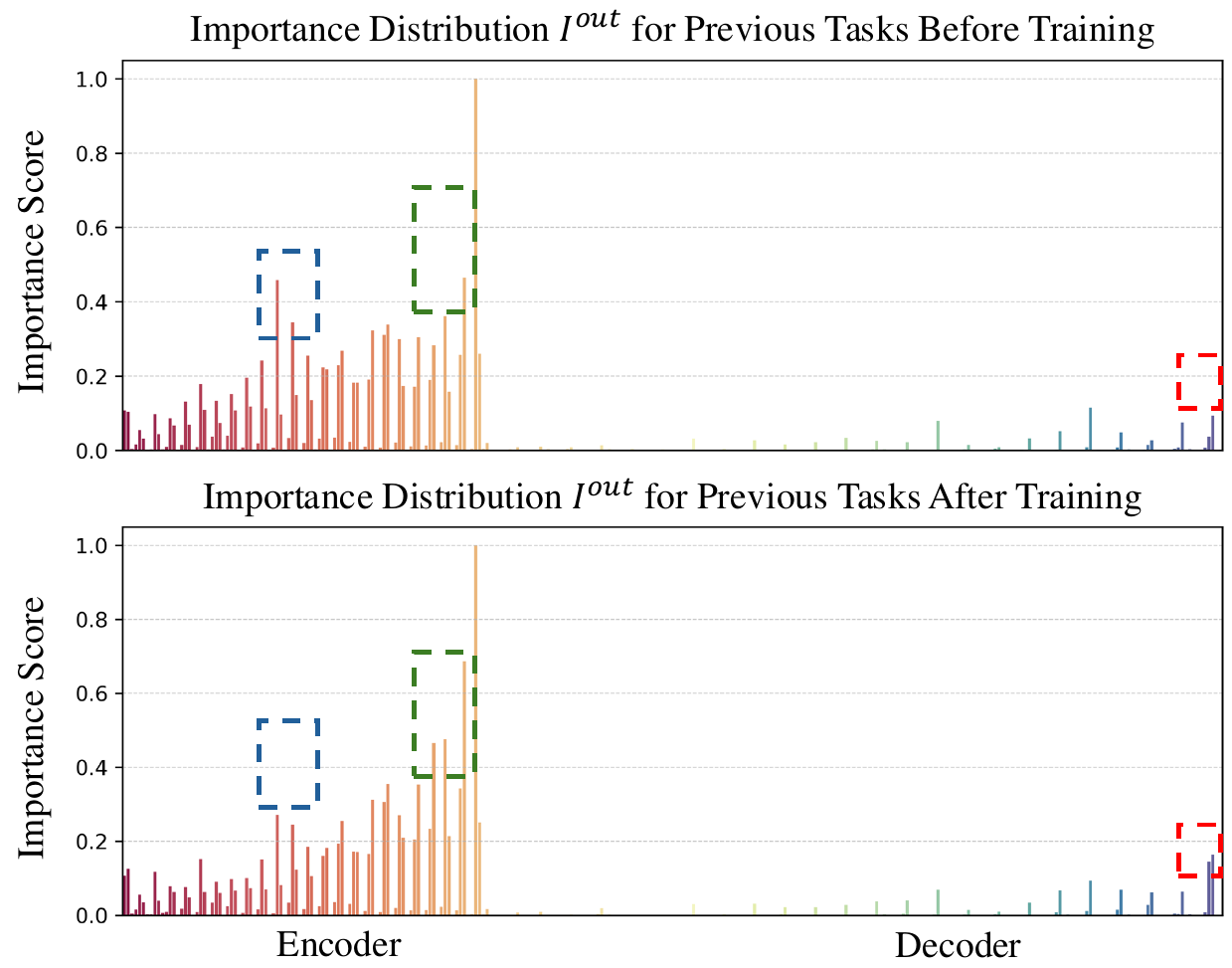}}
  \caption{Visualizations of task vector and parameter importance distributions on T5-large.}
  \label{fig:visualization}
\end{figure*}

\subsection{Ablation Study}
We conduct ablation studies to assess the effectiveness of the proposed techniques in {\ouralg}. The results for task order 1 on the Long Sequence Benchmark are shown in Table \ref{tbl:ablation}. Additional experiments, such as time complexity analysis, the impact of memory size, and hyperparameter sensitivity, are provided in Appendix~\ref{sec:hyper}.

\paragraph{Effect of Dynamic Importance Estimations.}  
To validate the role of dynamic importance estimation, we replace it with a static version (``- DIE''), where importance scores for historical tasks remain fixed after their initial computation.
The significant performance decline (3.1\% on OP and 1.4\% on BWT) highlights the necessity of dynamically updating historical task importance distributions. By maintaining up-to-date importance scores, our approach improves both robustness and accuracy, thereby enhancing knowledge retention and transfer.


\paragraph{Effect of Importance-Based Binary Mask Strategy in Knowledge Fusion.}  
We replace our knowledge fusion mechanism with three alternative model merging strategies:
(i) Without knowledge identification (``- KI''): Directly merge \(\tau^{\text{in}}\) and \(\tau^{\text{out}}\) in Eq.(\ref{eq:fusion}) without applying importance-based fusion.  
(ii) Global importance-based merging (``+ GM''): Use a global weighted sum of importance scores \(I_b^{\text{in}}\) and \(I_b^{\text{out}}\) for fusion instead of applying element-wise masking.
(iii) Adaptive Fusion \cite{yangadamerging} (``+ Adaptive''): A soft-masking method that uses raw importance scores directly for fusion instead of binary masks.
Additionally, we evaluate the effectiveness of updating the task-shared region by introducing ``- Share'', where $m^{in}$ is set to 0 when both $m^{in}$ and $m^{out}$ are 1. 

The results in Table \ref{tbl:ablation} confirm the effectiveness of importance-based binary masking in filtering redundant information and preserving task-specific knowledge.
Moreover, disabling updates to the task-shared region leads to performance drops of 2.1\% and 0.9\% on two evaluation metrics, demonstrating that updating task-shared parameters is critical for effective knowledge transfer between tasks.

\paragraph{Effect of Multi-Round Knowledge Fusion.}  
We compare multi-round fusion with traditional single-step fusion methods and analyze the impact of the number of knowledge fusions by adjusting the inner loop iteration size $Q$.
In {\ouralg}, with the total number of iterations for the inner loop fixed at \(N'\), increasing $Q$ reduces the number of fusion steps, which is $N'/Q$. A detailed analysis and the model’s time complexity are provided in Appendix~\ref{sec:time}.


Figure~\ref{fig:fusion} presents two key findings: (i) Multi-round fusion consistently outperforms single-step fusion by leveraging intermediate training information, resulting in smoother knowledge integration, similar to the distinction between gradient descent and stochastic gradient descent; and (ii) performance improves with the number of fusion steps up to a certain point, after which diminishing returns are observed.
This decline occurs because excessive fusion introduces noise and redundant updates, disrupting the balance between new and prior knowledge, and potentially causing overfitting to unconverged intermediate states. 


\begin{figure}[t]
  \centering
  \includegraphics[width=0.8\linewidth]{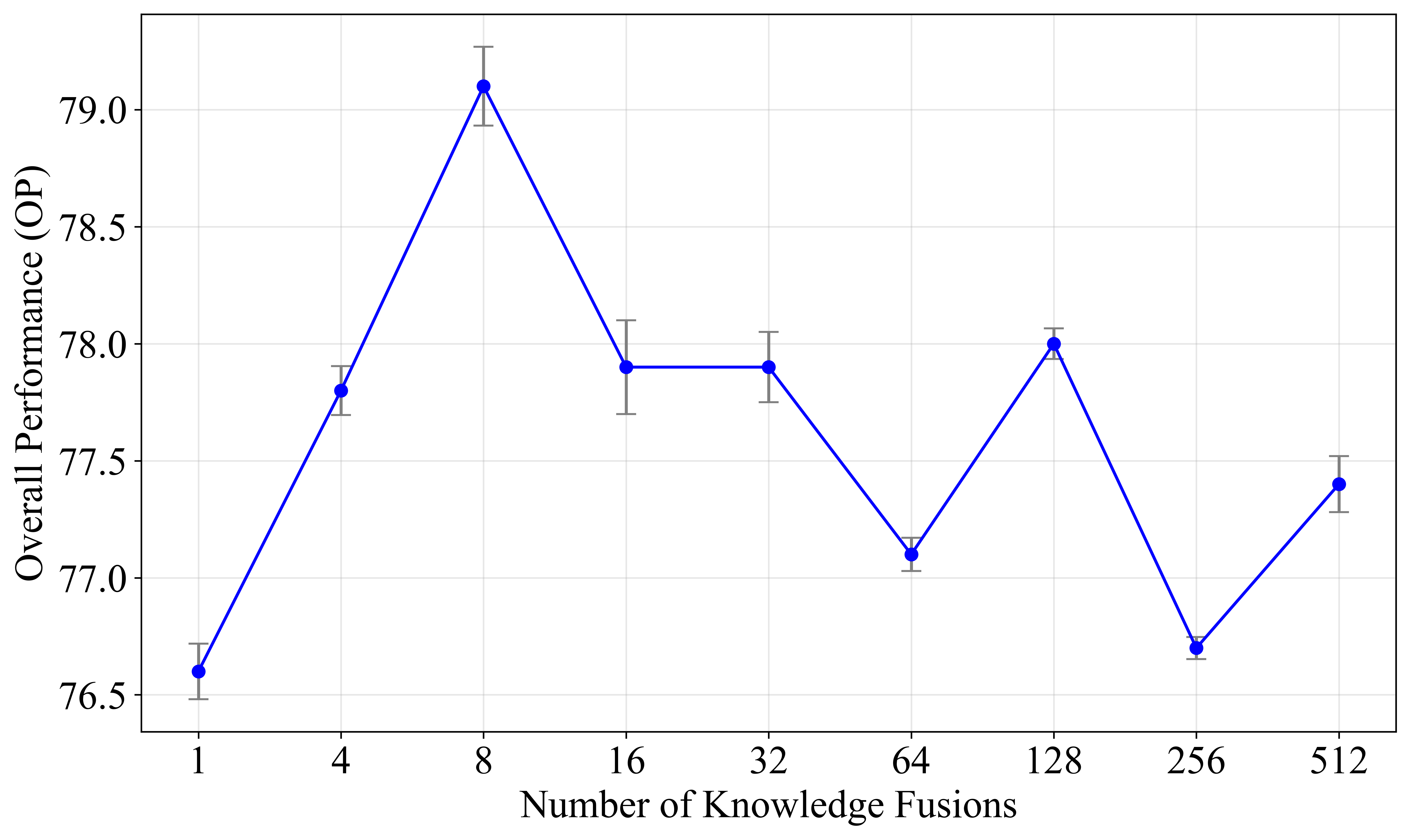}
  \caption{Ablation study on the number of fusions.}
  \label{fig:fusion}
\end{figure}

\subsection{Visualization}
We present two key visualizations to analyze the effectiveness of our proposed methods:
\paragraph{Can the Magnitude of the Task Vector Reflect Parameter Importance?}
We explore the relationship between task vector magnitude and parameter importance scores. As shown in Figure \ref{fig:visualization}(a), although the magnitude of parameter updates is generally large, only a subset of parameters, mainly in the encoder, are truly important.
This indicates that a large portion of the parameters are redundant, highlighting the need for our importance-based knowledge fusion mechanism.


\paragraph{Does the Importance Distribution of Historical Tasks Change with Model State?}
Figure \ref{fig:visualization}(b) illustrates the shift in the importance distribution of historical tasks before and after training on a new task. 
While the overall distribution remains stable across model states, notable changes in specific importance scores are observed, highlighted by the dashed box. This demonstrates the value of dynamic estimation, enabling more precise identification of key parameters and enhancing knowledge fusion across tasks.
A more detail analysis is provided in Appendix \ref{sec:preliminary}.

\section{Conclusion}



In this paper, we introduce Recurrent Knowledge Identification and Fusion ({\ouralg}), a novel CL framework that dynamically estimates the importance of parameters for previous tasks. {\ouralg} iteratively employs an inner learner to localize new knowledge and an outer learner to manage the global fusion of knowledge, enabling real-time and adaptive adjustments to the fusion strategy based on evolving importance distributions. Extensive experiments demonstrate the effectiveness of {\ouralg} in addressing continual learning challenges.


\section*{Limitations}

We acknowledge two limitations in this work.
Firstly, {\ouralg} is a rehearsal-based method. The outer loop relies on memory data to retrieve and dynamically update the parameter importance distributions of historical tasks.
This reliance may limit its applicability in scenarios where privacy concerns or data retention restrictions are present. Generative replay techniques could provide a solution by simulating the distribution of previous tasks without direct access to historical data.

Secondly, the time complexity of {\ouralg} increases with larger backbone models, primarily due to element-wise operations and multi-round fusion. For element-wise operations, global merging strategies have proven suboptimal, highlighting the need for balanced fusion granularity. Future work could explore focusing on specific important layers or adopting modular approaches to enhance efficiency. 
For multi-round fusion, we could further investigate how fusion frequency impacts performance and analyze the semantic knowledge learned at different stages of the training process. This could help minimize unnecessary iterations, while still preserving the benefits of iterative integration.

\section*{Acknowledgements}
We thank the anonymous reviewers for their valuable feedback and the support of the PolyU Research Student Attachment Programme.

\bibliography{acl2023}
\bibliographystyle{acl_natbib}

\appendix
\label{sec:appendix}
\section{Limitations of Static Parameter Importance Estimation}
\label{sec:preliminary}
\begin{figure}[t]
  \centering
  \includegraphics[width=1\linewidth]{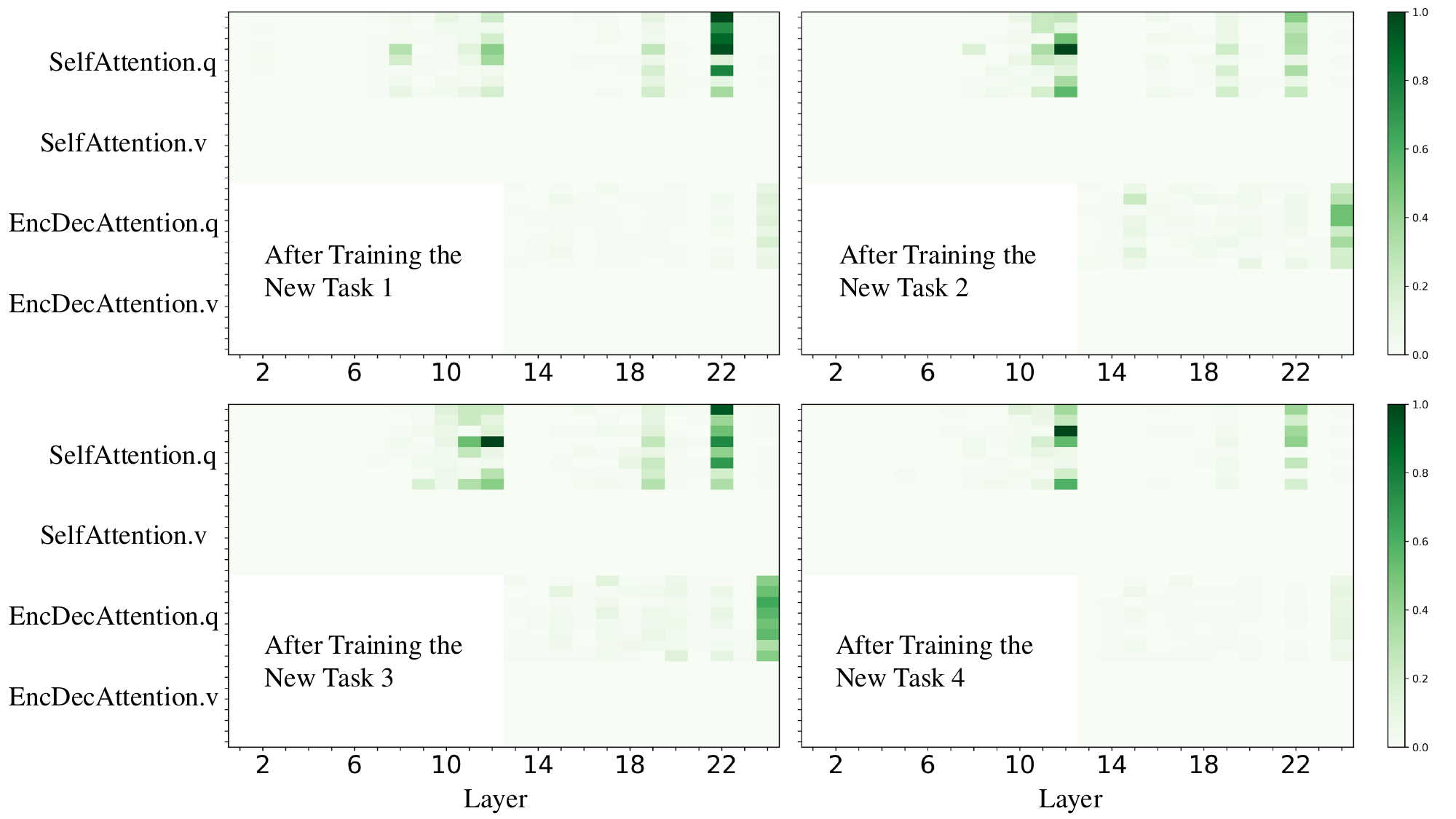}
  \caption{Visualization of the parameter importance distribution for the fixed historical task ``AG News'' after training on different new tasks.
  }
  \label{fig:heatmap1}
\end{figure}

In this section, we empirically demonstrate the limitations of static parameter importance estimation by analyzing how historical parameter importance distributions change under different conditions. Static importance estimation assumes that the importance scores of historical tasks remain fixed, which is both inaccurate and introduces biases during the knowledge fusion process. To highlight the dynamic nature of parameter importance, we analyze the following two aspects:

\paragraph{Changes in Historical Importance After Learning Different New Tasks.}
We investigate how the parameter importance distribution for a fixed historical task changes after the model is fine-tuned on different new tasks. Specifically, the model is trained sequentially on various new tasks, and the importance scores of the same historical task are re-evaluated using the updated model parameters.
As shown in Figure \ref{fig:heatmap1}, although the regions identified as important for the historical task remain largely consistent after learning different new tasks, the specific importance values exhibit noticeable differences.
This variability demonstrates that historical parameter importance is heavily influenced by the specific characteristics of the new task, making static importance estimation insufficient for accurately capturing the evolving model dynamics.


\begin{figure}[t]
  \centering
  \includegraphics[width=1\linewidth]{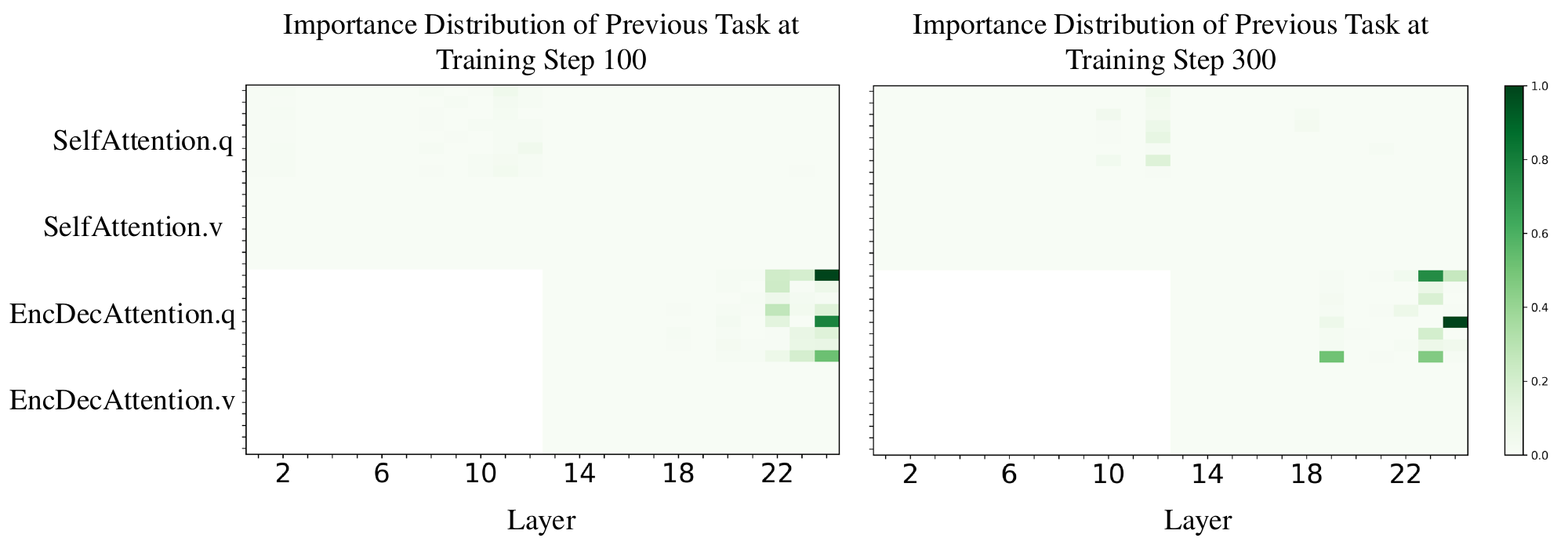}
  \caption{Visualization of the parameter importance distribution for a fixed historical task at different stages of training on a new task.
  }
  \label{fig:heatmap2}
\end{figure}

\paragraph{Temporal Changes in Historical Importance During New Task Training.} 
We further analyze how the historical parameter importance distribution evolves during the training process of a single new task. As shown in Figure \ref{fig:heatmap2}, by periodically evaluating the importance scores of a fixed historical task at different stages of training, we also observe temporal changes in the parameter importance distribution. These changes indicate that historical importance is not static even within the training process of a single task, reflecting the continuous interactions between new and historical knowledge. Static estimation fails to capture these temporal dynamics, which can lead to suboptimal knowledge fusion.

Our analysis confirm that historical parameter importance distributions are dynamic, influenced by both the characteristics of the new task being learned and the training stage. These observations provide strong empirical evidence supporting the need for dynamic importance estimation approaches.
Static importance estimation fails to account for these variations, potentially causing biases and inaccuracies in knowledge fusion. In contrast, dynamic importance estimation, as proposed in our framework, addresses these issues by continuously updating importance distributions to align with the most recent model state, ensuring more effective and accurate knowledge integration.


\section{Additional Results}
\label{sec:hyper}

\subsection{Effect on the SuperNI Benchmark}
To further validate the effectiveness of {\ouralg} in more complex CL scenarios, we have conducted additional experiments on the SuperNI Benchmark \cite{wang2022super}, which includes tasks like dialogue generation, information extraction, question answering, and summarization. Using T5-large as the backbone, we assessed 15 tasks following the experimental setup in \cite{zhao2024sapt}. The results are shown in the table \ref{tbl:SuperNI}.

\begin{table}[h]
\centering
\scalebox{1}{
\begin{tabular}{lcc}
\toprule
Method &   OP  & BWT \\
\midrule
\rule{0pt}{4pt} Replay & 35.4 & -15.8 \\
\rule{0pt}{8pt} O-LoRA  & 25.9 & -24.6\\
\rule{0pt}{8pt} TaSL  & 41.3 &   -12.7\\
\rule{0pt}{8pt} VR-MCL  &  40.5 & -10.9 \\
\rule{0pt}{8pt} {\ouralg}  &  43.3 & -8.4 \\
\rule{0pt}{8pt} MTL (Upper Bound)   &  50.7 & - \\

\bottomrule
\end{tabular}}
\caption{Overall results on the SuperNI Benchmark.}
\label{tbl:SuperNI}
\end{table}

While performance decreases due to task complexity, {\ouralg} consistently outperforms other methods, demonstrating its robustness and ability to handle more sophisticated CL scenarios.

\subsection{Effect of the Memory Size}
We investigate the impact of varying memory size on the performance of LoRAReplay and {\ouralg}. By adjusting the memory size per task \(|M|\) to {2\%, 5\%, 10\%, 50\%}, the results are shown in Table \ref{tbl:ablation_memory}. As expected, increasing the memory size generally improves the performance of all methods. 
{\ouralg} leverages its knowledge fusion mechanism to effectively preserve the parameters that store historical knowledge, thereby achieving better performance than LoRAReplay.

\begin{table}[h]
\centering
\scalebox{1}{
\begin{tabular}{lcccc}
\toprule
\multirow{2}*{\tabincell{c}{ }} & \multicolumn{4}{c}{Memory Size}\\
\cmidrule(lr){2-5}
 & 2\% & 5\% & 10\% & 50\%\\
\midrule
\rule{0pt}{6pt} LoRAReplay & 71.2 & 72.4& 73.8 & 76.1   \\

\rule{0pt}{8pt} {\ouralg} & 77.9 & 78.7  & 79.8 & 80.5 \\

\bottomrule
\end{tabular}}
\caption{Ablation study on memory size, using T5-large as the backbone.
}
\label{tbl:ablation_memory}
\end{table}

\subsection{Effect of Different LoRA Adapters}
We further investigate which components within a transformer block should incorporate LoRA. A typical transformer block consists of the query, key, and value (QKV) linear layers, the output linear layer (O) in the multi-head attention module, and the two linear layers in the feedforward network (FFN). Our analysis, presented in Table \ref{tbl:ablation_lora}, shows that applying LoRA to all these linear layers yields the best overall performance. Notably, adding LoRA to the FFN layers results in better BWT performance than applying it to the multi-head attention layers.

\begin{table}[h]
\centering
\scalebox{1}{
\begin{tabular}{lrr}
\toprule
LoRA Target Modules &   OP   & BWT \\
\midrule
\rule{0pt}{4pt} Attention Q V  & 77.9  & -3.4  \\

\rule{0pt}{8pt} Attention Q K V O & 77.7  & -3.3	   \\
\rule{0pt}{8pt} FFN  & 77.7   & -2.5	   \\
\rule{0pt}{8pt} Attention All + FFN & 78.0  & 	-3.2   \\

\bottomrule
\end{tabular}}
\caption{Ablation study on LoRA target modules, using T5-large as the backbone.}
\label{tbl:ablation_lora}
\end{table}

\subsection{Sensitivity Analysis for Hyperparameters}
The proposed framework incorporates three key hyperparameters, including the smoothing factor $\alpha$ for computing importance scores in Equations (\ref{eq:I}) and (\ref{eq:out}), the threshold $\delta$ for determining the importance of parameters, and the number of inner and outer loop training steps. Our analysis aims to assess the impact of varying these hyperparameters on our method's performance, testing on the T5-large backbone model.

As evidenced in Table \ref{tbl:hp_alpha}, we determine that the optimal setting for $\alpha$ is 0.55. An $\alpha$ value too low results in a performance decline, indicating that the calculated importance scores are not accurate.

\begin{table}[h]
\centering
\scalebox{1}{
\begin{tabular}{lcc}
\toprule
$\alpha_1$,$\alpha_2$ &   OP  & BWT \\
\midrule
\rule{0pt}{4pt} 0.35 & 77.7 & -3.4 \\
\rule{0pt}{8pt} 0.55  & 78.1 & -2.8\\
\rule{0pt}{8pt} 0.85  & 77.9 &   -3.4\\
\rule{0pt}{8pt} 0.95  &  77.8 & -2.7 \\

\bottomrule
\end{tabular}}
\caption{Performance comparisons of {\ouralg} equipped with different $\alpha$.}
\label{tbl:hp_alpha}
\end{table}

Regarding the selection of the threshold for important parameters, Table \ref{tbl:hp_delta} below shows the model's performance with varying thresholds $\delta$ on T5-large. It can be seen that setting a high threshold (50\%) reduces model effectiveness by categorizing less significant parameters as important, which can contaminate historical knowledge and lead to forgetting. Conversely, a 1\% threshold still maintains strong performance owing to our effective knowledge fusion approach, which preserves task-specific knowledge and prevents forgetting. Considering the 28-law of diminishing returns, we opted for a 20\% threshold to distinguish between important and less significant parameters.

\begin{table}[h]
\centering
\scalebox{1}{
\begin{tabular}{lcc}
\toprule
Importance Thresholds $\delta$  &OP &  BWT\\
\midrule
\rule{0pt}{4pt}1\%  &  77.4 & -3.3  \\

\rule{0pt}{8pt}10\% &  77.8 &  -3.1\\
\rule{0pt}{8pt}20\% & 77.9  & -3.4 \\

\rule{0pt}{8pt}50\% & 77.6  &  -2.4 \\
\bottomrule
\end{tabular}}
\caption{Performance comparisons of {\ouralg} equipped with different $\delta$.}
\label{tbl:hp_delta}
\end{table}

Finally, we investigate the effect of varying the number of inner and outer loop training steps on model performance. As shown in Table \ref{tbl:inner}, increasing the number of iterations for both the inner and outer loops can lead to improved performance, particularly in terms of the BWT metric. However, we observe diminishing returns beyond a certain point. Specifically, the performance gain becomes less significant when the number of iterations exceeds 8 for the inner loop and 4 for the outer loop. This suggests that our framework reaches an optimal balance between computational efficiency and performance with a modest number of iterations.

\begin{table}[t]
\centering
\scalebox{1}{
\begin{tabular}{llcc}
\toprule
Inner Steps ($Q$) & Outer Steps  &OP &  BWT\\
\midrule
\rule{0pt}{4pt} 4 & 1 & 76.7  & -4.4 \\
\rule{0pt}{8pt} 8 & 1 & 77.7  & -3.2\\
\rule{0pt}{8pt} 8 & 4 & 77.9  & -3.4\\
\rule{0pt}{8pt} 16 & 4 & 76.8  & -5.4\\

\bottomrule
\end{tabular}}
\caption{Performance comparisons of {\ouralg} (using T5-large as
the backbone) equipped with different inner and outer loop training steps.}
\label{tbl:inner}
\end{table}

In conclusion, it is worth noting that there is a small performance difference observed when varying the hyperparameters. This suggests that the proposed {\ouralg} method exhibits robustness and is not highly sensitive to the choice of hyperparameters.

\subsection{Time Complexity Analysis}
\label{sec:time}

In this section, we discuss the time complexity issues introduced by the techniques used in {\ouralg}. The additional time complexity can be explained qualitatively: assuming the number of training iterations for vanilla training is \(N'\), we set the total number of iterations for the inner loop to \(N'\) as well. This ensures a fair comparison with traditional methods, while also minimizing the number of iterations in our approach. In this case, the total number of iterations for our method is \(N' + (N'/Q)\), where \(N'/Q\) corresponds to the iterations of the outer loop. Typically, \(N'\) is 1024, and with \(Q\) set to 8, this results in an additional 12\% increase in training time.

Regarding the time consumption of knowledge identification and fusion, the variables used in the knowledge identification phase are derived from the gradients produced during normal training, requiring no extra computation time, only additional space to store parameter importance information. The knowledge fusion phase involves only simple univariate calculations, as shown in Equation~(\ref{eq:fusion}). Therefore, the overall time complexity does not increase significantly.

Quantitatively, we compare the training time of our method with LoRA Replay, as shown in Figure~\ref{tbl:time}. Compared to traditional LoRA replay methods, the addition of knowledge identification and fusion does not significantly increase training time across different backbones. For instance, when using LLaMA2-13B as the backbone, adding knowledge identification and fusion results in a 1.37x increase in training time compared to the original setup. However, for smaller models like T5-large and T5-xl, the training time remains relatively consistent, with no significant impact observed from the inclusion of the reasoning components.



\begin{table}[t]
\centering
\scalebox{0.65}{
\begin{tabular}{lrrrr}
\toprule
Training Time \\ (Min/Epoch) & T5-large & FlanT5-XL & LLaMA2-7B & LLaMA2-13B  \\
\midrule

\rule{0pt}{4pt}LoRAReplay	   &1.4 & 1.4 & 4.5 & 6.6	 \\
\rule{0pt}{8pt}O-LoRA	   &1.4 & 1.4 & 4.5 & 6.7	 \\
\rule{0pt}{8pt}VR-MCL	   &1.5 & 1.8 & 6.0 & 10.2	 \\
\rule{0pt}{8pt}TaSL	   &1.4 & 1.4 & 4.6 & 6.7	 \\
\rule{0pt}{8pt}{\ouralg}   &1.4 & 1.6 & 5.5 & 9.1 \\

\bottomrule
\end{tabular}}
\caption{Training time comparison across backbones.}
\label{tbl:time}
\end{table}

\subsection{Effect of the Different Importance Metric in Knowledge Identification}
We compare two alternative importance scoring approaches with Eq. (\ref{eq:1}): 
(i) using absolute gradients~\cite{michel2019sixteen}, $\left|\nabla_{w_{ij}} \mathcal{L}\right|$, instead of the gradient-weight product; and (ii) removing exponential moving average, relying only on importance scores computed from a single batch.

As shown in Table \ref{tbl:ablation_ipt}, our method with exponential smoothing outperforms the alternatives, with performance drops of up to 2.0\% and 1.3\% without smoothing. Similarly, using absolute gradients leads to lower performance compared to the gradient-weight product, underscoring the effectiveness of our approach in enhancing knowledge identification and model performance.

\begin{table}[h]
\centering
\scalebox{0.9}{
\begin{tabular}{lcc}
\toprule
Method & OP &  BWT\\
\midrule
\rule{0pt}{6pt} $I\left( \cdot \right) = \left|\nabla_{w_{i j}} \mathcal{L}\right|$  & 74.5 & -5.2\\
\rule{0pt}{8pt}  $I\left( \cdot \right) = \left|w_{i j} \nabla_{w_{i j}} \mathcal{L}\right|$ & 75.9 & -4.7 \\

\rule{0pt}{8pt} {\ouralg} (ours) & \textbf{77.9} & \textbf{-3.4} \\

\bottomrule
\end{tabular}}
\caption{Ablation study. Evaluating the impact of different importance metrics on knowledge identification.
}
\label{tbl:ablation_ipt}
\end{table}

\section{Dataset Statistics}
\label{sec:dataset}
Table \ref{long-sequence} show details of the datasets we used for our experiments, along with their evaluation metrics \cite{wang2023orthogonal, feng2024continual, xu2023seqcare}. 
For the Long Sequence benchmark, this includes five tasks from the standard CL benchmark (AG News, Amazon reviews, Yelp reviews, DBpedia and Yahoo Answers), four from GLUE benchmark (MNLI, QQP, RTE, SST2), five from SuperGLUE benchmark (WiC, CB, COPA, MultiRC, BoolQ), and the IMDB movie reviews dataset.
We report 6 different task orders used for our experiments in Table \ref{order}.
Table \ref{prompt} shows prompts for different tasks. NLI denotes natural language inference \cite{lu2021getting}, including MNLI, RTE and CB. SC denotes sentiment analysis, including Amazon, Yelp, SST-2 and IMDB. TC denotes topic classification, including AG News, Dbpedia and Yahoo \cite{xu2025dearllm, chen2024entity}.

\begin{table*}[htbp]
\centering
\scalebox{0.8}{
\begin{tabular}{lllll}
\toprule
\textbf{Dataset name} & \textbf{Category} & \textbf{Task}             & \textbf{Domain}     & \textbf{Metric} \\ \midrule
1. Yelp               & CL Benchmark      & sentiment analysis        & Yelp reviews        & accuracy        \\
2. Amazon             & CL Benchmark      & sentiment analysis        & Amazon reviews      & accuracy        \\
3. DBpedia            & CL Benchmark      & topic classification      & Wikipedia           & accuracy        \\
4. Yahoo              & CL Benchmark      & topic classification      & Yahoo Q\&A          & accuracy        \\
5. AG News            & CL Benchmark      & topic classification      & news                & accuracy        \\
6. MNLI               & GLUE              & natural language
inference                       & various             & accuracy        \\
7. QQP                & GLUE              & paragraph detection       & Quora               & accuracy        \\
8. RTE                & GLUE              & natural language inference                       & news, Wikipedia     & accuracy        \\
9. SST-2              & GLUE              & sentiment analysis        & movie reviews       & accuracy        \\
10. WiC               & SuperGLUE         & word sense disambiguation & lexical databases   & accuracy        \\
11. CB                & SuperGLUE         & natural language
inference                       & various             & accuracy        \\
12. COPA              & SuperGLUE         & question and answering                        & blogs, encyclopedia & accuracy        \\
13. BoolQA            & SuperGLUE         & boolean question and answering                & Wikipedia           & accuracy        \\
14. MultiRC           & SuperGLUE         & question and answering                        & various             & accuracy        \\
15. IMDB              & SuperGLUE         & sentiment analysis        & movie reviews       & accuracy        \\ \bottomrule
\end{tabular}}
\caption{The details of 15 datasets used in our CL experiments. First five tasks
correspond to the standard CL benchmark, all other tasks are used in long-sequence experiments.
}
\label{long-sequence}
\end{table*}

\begin{table*}[h]
\centering
\begin{tabular}{lll}
\hline
\textbf{Order} & \textbf{Model} & \textbf{Task Sequence}                                                                                                                                \\ \hline
1              & T5, LLaMA      & dbpedia → amazon → yahoo → ag                                                                                                                         \\
2              & T5, LLaMA      & dbpedia → amazon → ag → yahoo                                                                                                                         \\
3              & T5, LLaMA      & yahoo → amazon → ag → dbpedia                                                                                                                         \\ \hline
4              & T5             & \begin{tabular}[c]{@{}l@{}}mnli → cb → wic → copa → qqp → boolqa → rte → imdb →\\ yelp → amazon → sst-2 → dbpedia → ag → multirc → yahoo\end{tabular} \\
5              & T5             & \begin{tabular}[c]{@{}l@{}}multirc → boolqa → wic → mnli → cb → copa → qqp → rte\\ → imdb → sst-2 → dbpedia → ag → yelp → amazon → yahoo\end{tabular} \\
6              & T5             & \begin{tabular}[c]{@{}l@{}}yelp → amazon → mnli → cb → copa → qqp → rte → imdb →\\ sst-2 → dbpedia → ag → yahoo → multirc → boolqa → wic\end{tabular} \\ \hline
\end{tabular}
\caption{Six different orders of task sequences used for continual learning experiments. Orders 1-3 correspond to the standard CL becnhmark adopted by prior works. Orders 4-6 are long-sequence orders spanning 15 tasks, following \cite{razdaibiedina2023progressive}.}
\label{order}
\end{table*}

\begin{table*}[h]
\centering
\begin{tabular}{cl}
\hline
\textbf{Task}                                                       & \multicolumn{1}{c}{\textbf{Prompts}}                                                                                                                                  \\ \hline
NLI                                                                 & \begin{tabular}[c]{@{}l@{}}What is the logical relationship between the "sentence 1" and the "sentence 2"? \\ Choose one from the option.\end{tabular}                \\ \hline
QQP                                                                 & \begin{tabular}[c]{@{}l@{}}Whether the "first sentence" and the "second sentence" have the same meaning? \\ Choose one from the option.\end{tabular}                  \\ \hline
\begin{tabular}[c]{@{}c@{}}SC\end{tabular}   & What is the sentiment of the following paragraph? Choose one from the option.                                                                                         \\ \hline
\begin{tabular}[c]{@{}c@{}}TC\end{tabular} & What is the topic of the following paragraph? Choose one from the option.                                                                                             \\ \hline
BoolQA                                                              & \begin{tabular}[c]{@{}l@{}}According to the following passage, is the question true or false? Choose one \\ from the option.\end{tabular}                             \\ \hline
MultiRC                                                             & \begin{tabular}[c]{@{}l@{}}According to the following passage and question, is the candidate answer true \\ or false? Choose one from the option.\end{tabular}        \\ \hline
WiC                                                                 & \begin{tabular}[c]{@{}l@{}}Given a word and two sentences, whether the word is used with the same sense \\ in both sentence? Choose one from the option.\end{tabular} \\ \hline
\end{tabular}
\caption{Instructions for different tasks.}
\label{prompt}
\end{table*}

\section{Implementation Details}
\label{sec:details}
Experiments are implemented using PyTorch and the Transformer library, running on a single NVIDIA A100 GPU with 80GB memory. The following hyperparameters are used:

\begin{itemize}[leftmargin=*,itemsep=2pt,topsep=0pt,parsep=0pt]
\item \textbf{T5-large (770M)} and \textbf{FLAN-T5-XL (3B)}: 
Learning rate of 3e-4 for both loops, inner and outer batch sizes of 8, max input length 512, max target length 128, and 10 epochs. LoRA settings: $r = 8$, $\alpha = 32$, dropout = 0.05, targeting modules [q,v]. Testing: max new tokens = 128.

\item \textbf{LLaMA-2 (7B)} and \textbf{LLaMA-2 (13B)}:
Learning rate of 3e-4 for both loops, inner and outer batch sizes of 64, cutoff length 512, and 10 epochs. LoRA settings: $r = 8$, $\alpha = 32$, dropout = 0.05, targeting modules [q\_proj,v\_proj]. Testing: temperature = 0.02, top\_p = 0, top\_k = 1, num\_beams = 1, max new tokens = 128.

\end{itemize}

It is worth noting that we used the same hyperparameters across different datasets and backbones, demonstrating the generalizability of our method without requiring extensive hyperparameter tuning for each specific setting.

\section{Algorithm}
\label{sec:alg}
In this section, we provide the detailed implementation of {\ouralg} algorithm (see Algorithm~\ref{alg:my_algorithm}).

\begin{algorithm*}[t]
\caption{The Algorithm of the proposed {\ouralg}}\label{alg:my_algorithm}
\begin{algorithmic}[1]
\renewcommand{\algorithmicrequire}{\textbf{Input:}}
\renewcommand{\algorithmicensure}{\textbf{Output:}}
\REQUIRE Current task dataset $\mathcal{D}_k$, memory buffer $\mathcal{M}_{<k}$, model weights $\theta$, initial inner-loop learning rate $\beta_{in}$, outer-loop learning rate $\beta_{out}$, number of inner-loop steps $Q$, number of outer-loop steps $S$, hyperparameters $\alpha_1, \alpha_2$, total number of fusion steps $N$.

\STATE \textit{\# training iterations.}
\FOR {$b$ = $1,\ldots,N$}
    \STATE sample training data $\epsilon_{k}^q$ ($q = 1,\ldots,Q$) from $\mathcal{D}_k$
    \STATE $\theta_{b(0)} = \theta_b$
    \STATE \textit{\# inner loop.}
    \FOR{$q = 1$ \TO $Q$}
        \STATE obtain batch samples $\epsilon_{k}^q$
        \STATE $\theta_{b(q)}=\theta_{b(q-1)}-\beta_{in} \nabla \mathcal{L}\left(\theta_{b(q-1)}\right)$
        \STATE \textit{\# knowledge identification.}
        \STATE compute the importance $\bar{I}\left(w_{i j}\right)$ via Eq. (\ref{eq:1});
        \STATE update $I_{b(q)}$ via Eq. (\ref{eq:I})
        
    \ENDFOR
    \STATE calculate $\tau_b^{\text{in}} = \theta_{b(Q)} - \theta_{b(0)}$ and obtain $I^{in}_b$
    \STATE \textit{\# outer loop.}
    \STATE $\theta_{b(M)}^0 = \theta_{b(Q)}$
    \FOR{$s = 1$ \textbf{to} $S$}  
        \STATE sample memory data $\phi_{b}^s$ from $\mathcal{M}_{<k}$
        \STATE $\theta_{b(M)}^s = \theta_{b(M)}^{s-1} - \beta_{out} \nabla \mathcal{L}\left(\theta_{b(M)}^{s-1}\right)$
    \ENDFOR

    \STATE calculate $\tau_b^{\text{out}} = \theta_{b(M)}^S - \theta_{b(Q)}$ 
    \STATE calculate $I_b^{\text{out}}$ via Eq. (\ref{eq:out})
    \STATE \textit{\# knowledge fusion.}
    \STATE obtain $m_b^{\text{in}}$, $m_b^{\text{out}}$ via Eq. (\ref{eq:mask})
    \STATE update $\theta_{b+1} = \theta_b + (m_b^{\text{in}} \odot \tau_b^{\text{in}} + m_b^{\text{out}} \odot \tau_b^{\text{out}})$
\ENDFOR
\end{algorithmic} 
\end{algorithm*}

\end{document}